\def\year{2020}\relax
\documentclass[letterpaper]{article} 
\usepackage{aaai20}  
\usepackage{times}  
\usepackage{helvet} 
\usepackage{courier}  
\usepackage[hyphens]{url}  
\usepackage{graphicx} 
\usepackage{graphicx}  
\usepackage{multirow}
\usepackage{graphicx}
\usepackage{xcolor}
\usepackage{amsmath}
\usepackage{amssymb}
\usepackage{fdsymbol}

\makeatletter
\newif\if@restonecol
\makeatother
\usepackage[ruled,vlined]{algorithm2e}
\usepackage{tikz}

\usepackage{booktabs}

\DeclareMathOperator*{\argmax}{arg\,max}
\DeclareMathOperator*{\argmin}{arg\,min}

\title{Political Depolarization of News Articles \\ Using Attribute-aware Word Embeddings}

\author{Ruibo Liu,\textsuperscript{\rm 1} Lili Wang,\textsuperscript{\rm 1} Chenyan Jia,\textsuperscript{\rm 2} Soroush Vosoughi\textsuperscript{\rm 1}\\ 
\textsuperscript{\rm 1}Department of Computer Science, Dartmouth College\\ 

\textsuperscript{\rm 2}Moody College of Communication, The University of Texas at Austin\\ 

ruibo.liu.gr@dartmouth.edu, soroush.vosoughi@dartmouth.edu \\
}

\begin{document}


\maketitle

\linespread{0.93}
\begin{abstract}
Political polarization in the US is on the rise. This polarization negatively affects the public sphere by contributing to the creation of ideological echo chambers. In this paper, we focus on addressing one of the factors that contributes to this polarity, polarized media. We introduce a framework for depolarizing news articles. Given an article on a certain topic with a particular ideological slant (eg., liberal or conservative), the framework first detects polar language in the article and then generates a new article with the polar language replaced with neutral expressions. To detect polar words, we train a multi-attribute-aware word embedding model that is aware of ideology and topics on 360k full-length media articles. Then, for text generation, we propose a new algorithm called Text Annealing Depolarization Algorithm (TADA). TADA retrieves neutral expressions from the word embedding model that not only decrease ideological polarity but also preserve the original argument of the text, while maintaining grammatical correctness. We evaluate our framework by comparing the depolarized output of our model in two modes, fully-automatic and semi-automatic, on 99 stories spanning 11 topics. Based on feedback from 161 human testers, our framework successfully depolarized 90.1\% of paragraphs in semi-automatic mode and 78.3\% of paragraphs in fully-automatic mode. Furthermore, 81.2\% of the testers agree that the non-polar content information is well-preserved and 79\% agree that depolarization does not harm semantic correctness when they compare the original text and the depolarized text. Our work shows that data-driven methods can help to locate political polarity and aid in the depolarization of articles.

\end{abstract}

\section{Introduction}
\label{sec:introduction}

Political polarization refers to an individual's stance on a given issue that is heavily affected by their identification with a particular political party (e.g., Democratic or Republican) or ideology (e.g., liberal or conservative). Political polarization has been recognized as a major issue undermining western democracies~\cite{Youngmannb,groseclose2005measure,gentzkow2010drives}. Partisan media often exacerbates this problem; in fact, it is reported that politically polarized articles can gradually change the attitudes of communities~\cite{Iyyer}, limit the topics in public discussions~\cite{Chen2019a}, or even have a substantial effect on the evolution of policy~\cite{dardis2008media}. 

In this paper, we propose a data-driven framework to detect and reduce political polarity in a given news article. The framework is intended to be used by authors to depolarize their articles while maintaining their message. The tool will not change the stance of an author; it will transform the text to say the same thing, but in a less polar way. To illustrate this point, consider an article talking about illegal immigrants. The polarity of that article can be changed by simply changing the phrase \textit{illegal immigrants} with \textit{undocumented immigrants}. The polarity can be changed the other way around by changing the former phrase with \textit{illegal aliens}. This change in polarity does not affect the message of the article; it just softens (in the case of the first example) or hardens (in the case of the second example) the tone. The polarity of an article can be transferred by changing a few polarity markers without changing the meaning of the text; these polarity markers can be explicit phrases (e.g., \textit{illegal immigrants}) or more subtle expressions (e.g., \textit{can't pay for health insurance} vs. \textit{can't prove coverage of health insurance}).

Our framework is composed of two main parts: a polarity detection mechanism based on attribute-aware word embeddings, and an algorithm for fully automatic or semi-automatic depolarization. The polarity detection model detects phrases and expressions that make an article polar. The model is context-aware (topic-aware), meaning that a phrase could be detected as polar in articles about a particular topic while not being detected as polar in other topics. The depolarization algorithm suggests changes to the polar language detected by the polarity detection model that would reduce the polarity of the article, while at the same time preserving the meaning, the message and the stance of the article (e.g., the algorithm would not change the word \textit{ban} to \textit{approve} as that would change the stance of the article). Furthermore, the depolarization algorithm attempts to keep the article contextually meaningful, cohesive, comprehensible and grammatically correct (something that most current transfer models do not do well).

Our method is inspired by recent works on encoding attribute-specific information into word embeddings ~\cite{subramanian2018multiple,Lample2019,Logeswaran2018}, and works on adversarial text generation ~\cite{Ren2019,Alzantot2019,Chen2018,Zhao2018a,Chen2017,singh2018sentiment}. In our case, we treat political ideology and topic as two separate attributes and train a multi-attribute-aware word embedding model. Our embeddings can not only capture the general context of words, but also the nuanced differences in their usage in specific contexts, such as different topics and ideological slants. 
Next, we propose an algorithm, called TADA, for generating depolarized (or neutral) text that can replace polar language (identified using the multi-attribute-aware word embedding model) with grammatically correct \emph{neutral} language (also extracted using the our multi-attribute-aware model). TADA can be used fully automatically, where a new depolarized article is fully generated by the tool, or semi-automatically, where users can decide which of the suggested changes to accept. We evaluate our models and algorithm qualitatively and quantitatively.

\section{Related Work}
\label{sec:related_work}
Political polarity and bias in the media have long been studied in the fields of communication and political science, going back to at least the 1950s ~\cite{scammell2018media}. Its impact on the political process has also been extensively studied~\cite{Kahneman1984,Author2009,glaeser2007corruption}. Specifically, studies show that if news consumers regularly receive polar news, they are prone to adopting similar polar views, and tend to start exclusively ``following" the media outlets whose reporting conforms with their established beliefs, thus spiralling into an echo chamber where their bias is further reinforced. 

\noindent \textbf{Polarity Detection} There is a wealth of work on how to detect polarity in media. Yano et al.~\cite{Yano} performed sentence-level manual polarity annotation on blog posts from 2008 and they showed that words capturing emotion and named entities of opposing political parties can be a strong indicator of polarity. 
In recent years, automated identification of political polarity and bias in news has gained more attention. Gentzkow et al.~\cite{gentzkow2010drives} derive a "slant index" to rate the ideological polarity of newspapers. They first identify words that are used much more frequently by one party than by another from the Congressional Record, and then compute a slant index (same concept as polarity) based on the frequency of these party-specific tokens. Lin et al.~\cite{lin2011more} proposed a scheme for bias categorization. The scheme includes the political party, frequently mentioned legislators, region, ideology, and gender. 
Iyyer et al.~\cite{Iyyer} apply a recursive neural network (RNN) to identify the political position evinced by a sentence. Owing to RNN's modelling power on sequential data, they claim that their model can capture both syntactic and semantic features. However, in terms of the detection output of the highest probability n-grams for two ideologies, their model struggles to distinguish non-polar content and polar indicators. Lexical variations have also been studied in non-political contexts. Shoemark et al. ~\cite{shoemark2017aye} conduct a large-scale study of dialect variation on Twitter in the UK. Their data-driven approach identifies Scotland-specific lexical features.

\noindent \textbf{Text Style Transfer} There are parallels between the depolarization stage of our framework and the general field of linguistic style transfer. One of the mainstream methods for style transfer is to rely on variants of auto-encoders to generate ``transferred" text~\cite{Zhou,Mueller2017,Guu,Yang2017,Bowman}. Works that rely on auto-encoders for style transfer require a large parallel corpus. There is a lack of ideal parallel corpus for our task that aligns sentences with the same content but different political ideology~\cite{subramanian2018multiple,tian2018structured}. Due to the lack of such corpus and also because of the difficulty of disentangling polar text with non-polar content~\cite{Dai,prabhumoye2018style}, these works fail to output meaningful indicators of polar text and in most cases, the output is problematic in grammar and hardly understandable ~\cite{wu2019hierarchical}. Chen et al.'s ~\cite{Chen2019a} work, which uses style transfer to flip the bias in the news headlines, is the most similar to ours. They first analyze bias among the articles and then use a cross-aligned auto-encoder trained on opposite-ideology news data to generate flipped titles. The main limitation of this work, as stated by the authors, is that even in their successful cases, the overlap of generated and ground-truth headlines is very low, which means the auto-encoder model has the tendency to discard too much information even from non-polar content part. We differ from their work in the model selection and also integrating salient words up-sampling into the feature selection (this is explained in detail later in the paper). Also, we extend the task to the depolarization of the entire article, instead of only headlines, which is a more challenging problem since we have to preserve contextual and grammatical correctness across the whole article.

\begin{figure*}[ht]
\centering
\includegraphics[width=\textwidth]{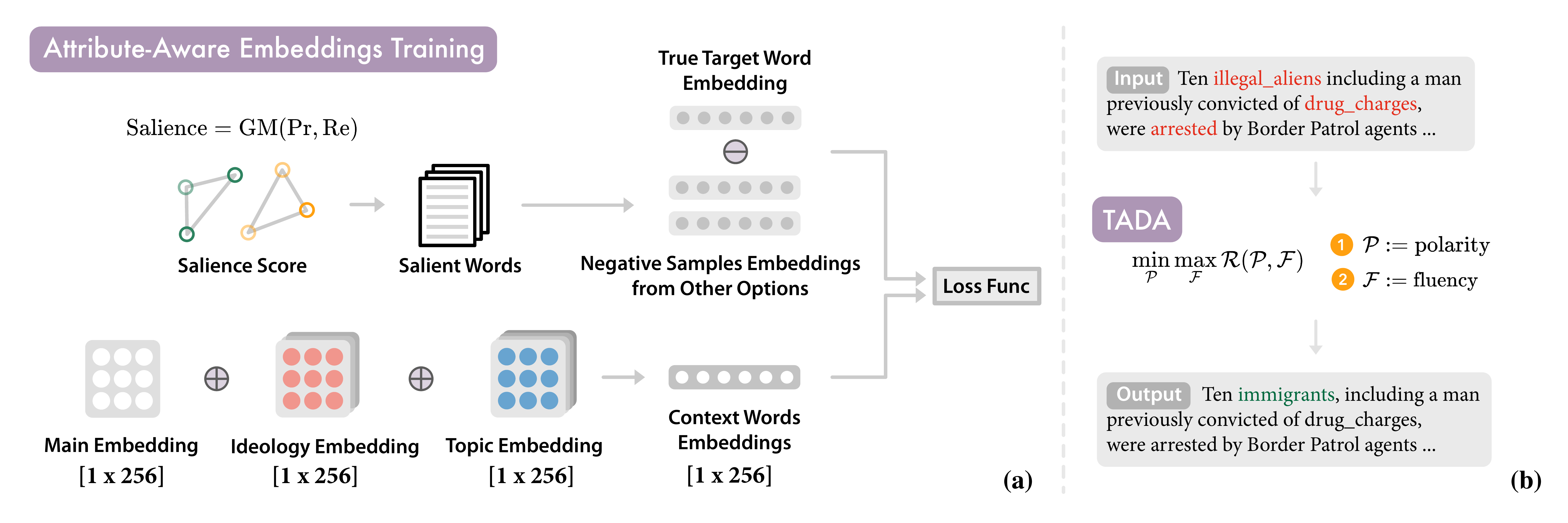}
\vspace{-0.25in}
\caption{Overview of (a) dynamic attribute-aware embeddings training, and (b) TADA algorithm of our framework.}
\vspace{-2.5mm}
\label{fig:emb}
\end{figure*}

More recent works on style transfer pay more attention to the quality of generation and whether the newly created text can successfully preserve the non-style content. These approaches are mostly template-based: they first identify target words that contribute a lot to style, and then replace these words and generate new expressions. Li et al.~\cite{Li2018a} propose a simple but efficient approach called ``Delete, Retrieve, Generate", which deletes stylized n-grams based on corpus-level statistics and stylizes it based on similar, retrieved sentences. Wu et al.'s work ``Point-Then-Operate (PTO)"~\cite{wu2019hierarchical} also follows this two-step manner to transfer style, which consists of a high-level agent that proposes operation positions and a low-level agent that alters the sentence. Huang et al.~\cite{Huang2019} bring synonyms from a dictionary as external reference information to improve generation quality, and they view the transfer generation task from the point of view of human's: humans tend to replace words or phrases in the original sentence with their corresponding synonyms and make necessary changes to ensure the new sentences are fluent and grammatically correct. Our work is inspired and builds upon these dictionary-guided methods, but rather than using an off-the-shelf dictionary that can only guarantee grammatical correctness, we train a ``dynamic dictionary" (dynamic with respect to topic and ideology) on an attributes-labeled corpus so that such a dictionary can not only preserve grammar but also can serve as a strong reference for a depolarized-text generation. This paper is related to another attempt from our lab at transferring the political polarity of news articles using a transformer-based conditional generator \cite{LiuCSCW2021}.

\section{Approach}

Our model is comprised of a dynamic multi-attribute embedding model which is used to detect polar phrases and a probabilistic algorithm for depolarizing the detected polar text. In this section we explain our framework in detail.

\subsection{Dynamic Attribute-aware Model}
\label{sub:model}
In order to capture attribute-invariant as well as multi-attribute-aware information, we extend a current single-attribute dynamic model \cite{Gillani:2019vi} to multi-attribute case through making following changes: (1) we adopt a two-step procedure, first we pre-train attribute-specific embeddings using datasets for each attribute (i.e., we train embedding models using articles with different ideologies and topics). Next, we jointly train am attribute-invariant embedding on the full dataset with attribute weights initialled as pre-trained embeddings, (2) we integrate salient words (words that can help distinguish one attribute option from another) features into the joint training step to promote saliency of attribute-specific words, (3) we replace the original negative sampling objective function with attribute-aware negative sampling objective so that our training can converge faster and the model is able to reflect attribute features in the final embeddings.

The basic idea of our attribute-aware model for the embedding of a word $w$ can be defined as follows:
\begin{multline}
    E(w, A_1,...,A_k)  = E_{m}(w) +  \sum_{A_1}p_{A_1} E_{A_1}(w) +... \\
    + \sum_{A_k}p_{A_k} E_{A_k}(w)
\end{multline}
where $E_{m} \in \mathbb{R}^{|V|\times \textrm{emb}}$ denotes the attribute-invariant embedding (main embedding) of $w$, and $E_{A_1}(w),..., E_{A_k}(w) \in \mathbb{R}^{|V|\times k}$ represents the shift of main embedding in specific attribute domain. We denote the embedding corresponding to each possible option for an attribute $A$ as $E_{A}(w)$; note that this framework supports embeddings for multiple attributes (denoted by $E_{A_1}(w)$ to $E_{A_k}(w)$). For each option in each attribute, we sum up all embeddings from all options in a weighted form ($p_{A_i}$ is the weight vector for attribute $i$), and add it to the main embedding. 
We add these embeddings along the embedding axis and use the new summed-up matrix ($E(w, A_1,...,A_k)$) as the final embedding. The resulting model is shown in Figure~\ref{fig:emb} (a).

\subsubsection{Integrating Salient Words}
\label{subsub:up_sample}

One limitation of our ``main + attribute" structure is that all negative samples that help training converge come from the same vocabulary set, meaning that we cannot have the model pay more attention to words that are more salient in one category (within the same attribute). This becomes an issue when an attribute has many categories (for instance our topic attribute which has 11 categories). As shown in Table~\ref{tab:before}, the most similar terms of the same query word \textit{illegal} show nearly no difference when we switch the topic. 

Certain words are strongly associated with a particular attribute. For instance the phrase  \textit{illegal\_aliens} is a strong indicator of conservative ideology, which means that it should be a good candidate of negative samples in the other two ideology options. Based on this observation, we identify a set of salient words that serve as strong indicators of different attribute options. We integrate these salient words into our training by making them more likely to appear in negative samples. This way our attribute-aware model can converge faster and output higher-quality word embeddings. Specifically, the salience score $\mathcal{S}$ for a word $w$ with respect to the attribute option $O_i$ is:

\begin{equation}
  \mathcal{S}(w, O_i) = \textrm{GM}(\frac{\textrm{count}(w, O_i)}{\sum\limits_{n=1}^{N} \textrm{count}(w, O_n)},\  \frac{\textrm{count}(w, O_i)}{\sum\limits_{w_i \in |V|} \textrm{count}(w_i, O_i)}) 
\end{equation}

\noindent where the first fraction is the probability that a word $w$ is from option $O_i$ (i.e., $\mathbb{P}(O_i|w))$, while the second fraction is the probability that $w$ can be found in $O_i$ (i.e., $\mathbb{P}(w|O_i)$). $\textrm{GM}$ is the geometry mean of these two terms. Note that we perform laplace smoothing and normal distribution normalization on the two terms.

Using this definition, the salience score of a word $w$ with respect to option $O_i$ depends on $\mathbb{P}(w|O_i)$ and $\mathbb{P}(O_i|w)$. This means that a word with a high salience score is highly frequent in $O_i$ while not being universally frequent (similar to the TF-IDF score). We use this score to sort our vocabulary set for each option, and then we apply a weighted (in terms of salient score) up-sampling on those words that are marked as salient (we explain this in more detail in the following section). Specifically, when we generate negative samples for a given activated option of an attribute (e.g., the option liberal for ideology attribute), we randomly pick words with a certain probability that has been modified by the salient words up-sampling algorithm.

\subsubsection{Attribute-aware Objective Function}
\label{subsub:obj_function}

We choose a CBOW (Continuous Bag of Words) model ~\cite{Mikolov} as our base to train the attribute embedding since it is reported to perform relatively well with large datasets~\cite{Schnabel2015}. The original negative sampling objective function from Mikolov's work ~\cite{Mikolov} can be represented as:

\begin{equation}
    J(\theta) = \argmax_\theta (\log\ \sigma(v_c\cdot v_w) + \sum_{(w, c) \in D'} \log\ \sigma(-v_c\cdot v_w))
\end{equation}

\noindent where $v_w$ and $v_c$ are a pair of word and context in vector form, $\sigma$ denotes the sigmoid function and $D'$ is the set of so-called ``negative samples". The goal is to find parameters to maximize the probability of positive cases (context words $c$ are indeed around word $w$) while minimize the probability of negative cases (word-context pairs $(w, c) \in D'$ where $D'$ are a set of randomly sampled pairs). This works well if no attribute is considered. However, it is highly possible that some negative cases we randomly sample for one attribute are actually positive samples in another attribute. As described earlier, this problem becomes even more severe as the number of attributes and the options for the attributes increases. 

In the previous section we described how we create a list of salient words for each attribute option. We integrate these words in the objective function to improve negative sampling and dampen the issues described above. We do this by creating a ``reversed" salient-word bank for each option $O_i$, which is in turn used to generate negative samples. Words that are salient with respect to an option appear less often in the reversed bank of that option and vice versa. The count for word $w$ in the reversed salient-word bank for option $O_i$ is calculated as:

\begin{equation}
    \textrm{Count}(w) = M \times \frac{1}{p}, \textrm{where}\ p = \frac{\mathcal{S}(w, O_i)}{\sum_{w \in V}\mathcal{S}(w, O_i)}
\end{equation}

\noindent here $\mathcal{S}(w, O_i)$ is the salience score we defined in Equation 2, $M$ is a relatively large number ($10^8$) close to the total count of words in our dataset. This equation ensures that words that are salient with respect to an attribute option will appear less in that options word bank and thus has less chance of being selected as a negative sample and vice versa. We generate and store these words banks in advance to speed training.  

Through this salient-based negative sampling method, we are able to improve the quality of our attribute-aware word embeddings. Our model is evaluated later in this paper.

\subsection{Detect and Depolarize Polar Text}
\label{sec:detect_and_depolar}
\subsubsection{Detect Polar Words}
Using the attribute-aware model, we define a measure of polarity for a word as the distance between its embeddings in different ideology attributes (liberal, neutral, conservative). Mathematically, our definition of polarity score $\mathcal{P}$ for each word $w$ in topic $t$ is:

\begin{equation}
    \mathcal{P}(w, t) = \sum_{s_i,s_j \in S} \cos(E(w, s_i, t), E(w, s_j, t))
\end{equation}
\noindent where $S$ is the set of all ideology attributes and $E$ is the attribute-aware embedding. We apply z-score normalization on each word's polarity score within the same topic. As a result all polarity scores' mean $\mu$ is located at zero, with several sigma $\sigma$ deviation in two directions. We treat the words that have above-zero polarity as polar words, and for the sake of generalization we filter out words whose frequency is lower than 500.

The polarity of a paragraph (or even a full article) can be calculated by adding the polarity score of each polar word (a word with normalized polarity greater than 0). We ignore non-polar words in our calculation as we do not want the polarity of a paragraph to be artificially lowered due to its length. At this point, we have the polarity score of each paragraph and a set of polar words in the paragraphs. The next step is to replace those polar words with neutral words such that the polarity of the paragraphs go down.

\subsubsection{Depolarized-Text Generation}
\label{sec:TADA}
In this section, we introduce the Text Annealing Depolarization Algorithm (TADA) as our solution to the depolarization task. The workflow of TADA is shown in Figure~\ref{fig:emb} (b).

In addition to polarity detection, the multi-attribute-aware word embedding model can also be used to find a polar word's neutral replacement. 

For a word $w_i$ from source ideology $s_i$ within topic $t$, its neutral expression $w_{\mathcal{N}}$ can be retrieved using the following equation:

\begin{equation}
    w_{\mathcal{N}} := \argmin(\cos(E(w_i, s_i, t), E(w_i, \textrm{neutral}, t)))
\end{equation}

\noindent this amounts to retrieving $w_i$'s most similar word in the neutral ideology and the same topic $t$. The distance is measured using cosine similarity. To ensure the correctness of grammar, we filter out those candidates whose part-of-speech (POS) are not the same as the original word's POS. For each detected polar word, a neutral candidate list (denoted as $w_{\mathcal{N}}$) is retrieved and sorted using this method. We limit the list for each polar word to the top 20 candidates for the search algorithm (which means $|w_{\mathcal{N}}| = 20$).

TADA is a modified simulated annealing algorithm that balances depolarization performance and fluency. The algorithm is also highly efficient. Let us first define the optimization goal of our task: Given a paragraph consisting of words $\overline{w}$ in topic $t$, assume after $s$ replacement steps we obtain a new sequence of words $\overline{w}^* = \{w_1,...,w_m\}$. The reward $\mathcal{R}$ of the whole modification is:

\begin{equation}
  \mathcal{R}(\overline{w}^*, \overline{w}, t) = \frac{1}{s + \lambda}[\underbrace{ \overbrace{-\mathcal{P}(\overline{w}^*, t)}^{\text{step 1}} + \mathcal{F}(\overline{w}^*, \overline{w})}_{\text{step 2}}]
  \label{reward_eq}
\end{equation}

\noindent where $\mathcal{P}$ is the polarity score we mentioned in the previous part, $\mathcal{F}$ is the fluency score of the newly generated sequence $\overline{w}^*$ with respect to the original paragraph $\overline{w}$. Also, we use $s$ to normalize the total rewards in terms of how many steps it takes to reach the optimum replacement, and we add a regularization term $\lambda$ to avoid infinity reward $\mathcal{R}$ caused by zero step/no modification (i.e., $s=0$). We set the hyperparameter $\lambda = 0.01$ through empirical observation.

As shown in Equation \ref{reward_eq}, we compute the reward in a two-step manner: in step one, we compute the polarity score $\mathcal{P}$ of the randomly generated modified sequences ($\overline{w}^*$) (note that a modification is only made when the polarity score of the sentence is lowered); and then in step two, we sum up the fluency score $\mathcal{F}$ and the negative polarity score of each candidate sequence to produce the final reward $\mathcal{R}$. Given such a reward definition, the training goal of our TADA algorithm is to pick the proper sequence modification that has the highest reward after several iterations. 

It is possible that no modifications will be made in the first step if there are none that reduce the polarity score. In that case the second step will not go into effect and the algorithm will terminate with no changes on the sequence. To decrease the number of unchanged cases we can set a lower $T_{\min}$ in Algorithm~\ref{alg:tada}, which leads to more iterations for the depolarization.

We choose the BLEU~\cite{papineni2002bleu} score as an estimation of fluency ($\mathcal{F}$). BLEU is computed based on how many n-gram overlap between the newly generated text and the original one. We believe it is more suitable than other language model scoring methods because the replacement (i.e., depolarization) causes a slight stylistic shift (words with the same meaning but different styles). Compared to other language model methods (e.g., BERT) BLEU is more sensitive to such changes. Another reason for selecting BLEU is speed. Some neural language models, like BERT, suffer from slow evaluation scoring~\cite{lan2019albert}, Since we are not rewriting the whole sentence but performing modification on the tokens, we choose BLEU as our light-weight and fast fluency metric.

\begin{algorithm}
\DontPrintSemicolon
\KwIn{Original polar text $\langle w_1, \ldots, w_m \rangle$}
\KwOut{Neutral text $neutral$}
$T \gets T_{\textrm{max}}$\;
$neutral \gets $[]\;

\For{$w_1$ \textbf{to} $w_m$} {
  \If{$\mathcal{P}(w_i) >$ \textrm{threshold} ($\mathcal{P}(w_i)$ by Eq. 5)}{
    retrieve $w_{\mathcal{N}}$ of $w_i$ (by Eq. 6)\;
    }
}
 $t \gets 0$\;
\While{$T>T_{\textrm{min}}$}{

    \textrm{Randomly init} $best$ \textrm{from} $w_{\mathcal{N}}$\;
    $next \gets $ \textbf{RandomPickCandidates($T, best$)}\;
    $\Delta \mathcal{R}\gets$ $\mathcal{R}$(\textrm{next}) $-$ $\mathcal{R}(\textrm{best})$ ($\mathcal{R}$ by Eq. 7)\;
    \uIf{$\Delta R < 0 \land random < \mathbb{P}_{\textrm{accept}}(T, \Delta \mathcal{R}$)}{
    $best\gets next$\;
    }
    \Else{
    Discard $next$\;
    }
    
  $t \gets t + 1$\;
  $T \gets \frac{T_{\textrm{max}}}{\log(t+d)}$\;
}
$neutral \gets$ \textrm{Merge} $best$ back to text\; 
\Return{neutral}\;
\caption{Text Annealing Depolarization}
\label{alg:tada}
\end{algorithm}

Given the reward function, the fully-automatic mode relies on an efficient search algorithm that can find the best combination of neutral words. Given an article whose length is $l$ and the vocab size is $|V|$ (a very large number), the search space of common sequential decoding methods~\cite{shen2017style,john2018disentangled,dai2019style} is $|V|^l$, and the final generation often drops into a local optimum because of its accumulation feature. Reinforcement-learning-based methods ~\cite{gong2019reinforcement,xu2018unpaired} have more accurate quality control over the generation and partially overcome the local optimum problem, however, these methods rely on continuous sampling of candidate sequences (still $|V|^l$ search space),  which result in low efficiency. Our algorithm, TADA, improves the efficiency in two perspectives: We first narrow the search space for each token from $|V|$ to $|w_{\mathcal{N}}|$, which is a limited-length neutral replacement words list proposed by our dynamic attribute-aware model. Then, instead of using sequential decoding to generate a replaced sequence, we use the simulated annealing algorithm as our template, and modify it to meet our requirements. The overview of our algorithm (called TADA) is shown in Algorithm~\ref{alg:tada}.

We keep the general framework of the classical simulated annealing algorithm: The optimization procedure begins with a maximum temperature $T_{\max}$, and after several iterations the optimization stops, when the temperature reaches $T_{\min}$. Instead of relying on the normal token-by-token decision, we decouple the search space with the length of the input. The replacement only happens when there are polar words detected, and we deploy a propose-and-judge manner as the core logic of TADA. In this way, assuming there are $k$ words being detected as polar words, the search space of our algorithm is thus ${|w_{\mathcal{N}}|}^k$, where $k$ is not necessarily related to the input length and $|w_{\mathcal{N}}|$ is limited to 20. We make use of the efficient search method in the classical simulated annealing algorithm. Each iteration of our algorithm can be described in three steps:

\begin{enumerate}
    \item Randomly pick polar words in a paragraph, and randomly pick neutral candidates for each selected polar word. Use these candidates to do the replacement, and compute the reward score ($\mathcal{R}$).
    \item If the newly computed reward score is smaller than the best, which means the polarity has decreased and fluency is well-conserved, accept these candidates to generate a new paragraph.
    \item If not, accept the candidates with probability $\mathbb{P}_{\textrm{accept}} = \exp(-\frac{\Delta \mathcal{R}}{T})$.
\end{enumerate}

We also use several tricks to improve the generation quality. For example, in order to guarantee that the picked candidates are grammatically correct, we filter out the candidates whose POS is not the same as the POS of the original word. Also, we dynamically adjust the ``cooling speed" ($\Delta T$) based on the length of the paragraph since longer paragraph may need more iterations to find the best candidates combination. In general, given the observation that certain polar words or phrases contribute a lot to the polarity, TADA adopts a propose-and-judge procedure, avoiding the unnecessary token-by-token sequential decoding from scratch. We make use of a pre-trained attribute-aware model to shrink the search space of neutral replacement to guarantee our algorithm runs efficiently. (All the parameters used in the model are listed in the Implementation section.)

\begin{figure*}[ht]
  \centering
  \includegraphics[width=\linewidth]{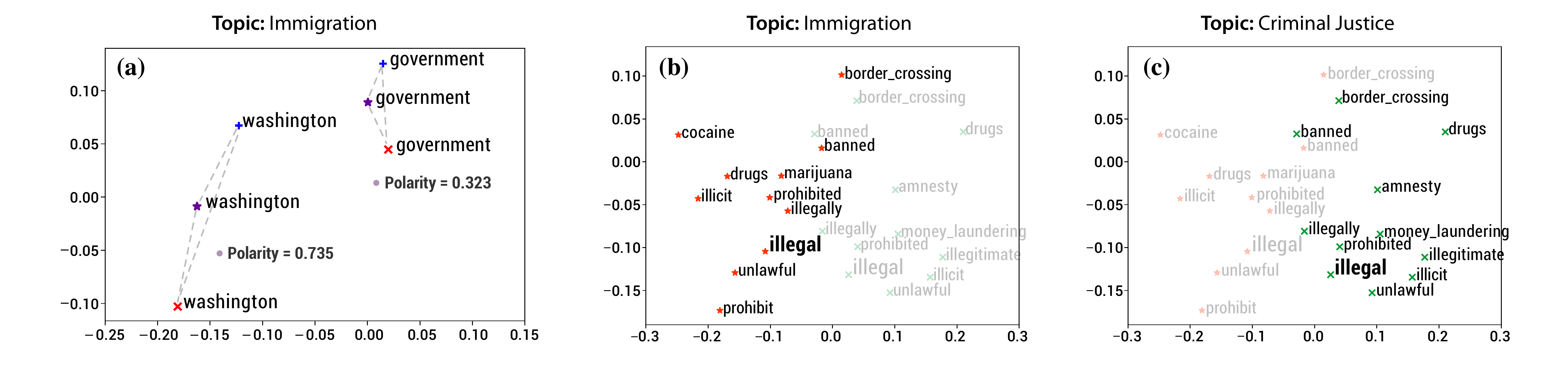}
  \vspace{-0.2in}
  \caption{Several visualization examples in embedding space to verify our model manages to capture attribute-aware polarity. In (a), we plot 2D projections of the words ``\textit{government}" and ``\textit{washington}" from liberal, neutral and conservative within the same topic ``Immigration". In (b) and (c), we plot the ten nearest words to the word ``illegal" in two different topics (``Immigration" and ``Criminal Justice"). We prove that our model can not only measure the lexical polarity within the same topic (ideology-aware), but also reflect the discrepancy between different topics (topic-aware).}
  \label{fig:diff}
\end{figure*}

\section{Implementation}
\subsection{Datasets}
The data for this project was collected from Media Cloud\footnote{https://mediacloud.org/}, which is an academic research project led by MIT and Harvard University. The platform collects articles from a large number of media outlets and makes the content available via an API. 

We followed the political issues defined by a survey-based website\footnote{https://www.isidewith.com/polls} and chose 11 topics (shown in Table~\ref{tab:polarity_score}) four our data collection. For each topic, we use the words under the corresponding title on isidewith.com as query keywords. (E.g. topic: Social Issue $\rightarrow$ query terms: \emph{abortion}, \emph{gay marriage}, \emph{death penalty}, etc) We collected around 360k full-length articles from May 1st, 2018 to May 1st, 2019 (about 6GB of plain text), as our training data (similar to Liu et al. \citeyear{liu2021mitigating}). The articles are from 22 media outlets. We assign an ideological polarity label to each outlet using data from the Pew Research Center. The data from Pew is based on a survey of news consumption by people with different political affiliation ~\footnote{https://www.journalism.org/interactives/media-polarization/table/overall/}. The labels we assigned for polarity are \textit{liberal} (corresponding to Pew's \textit{consistently liberal} and \textit{mostly liberal} labels), \textit{neutral} (corresponding to Pew's \textit{mixed} label), and \textit{conservative} (corresponding to Pew's \textit{consistently conservative} and \textit{mostly conservative} labels). Each article was also labeled with a topic. The assigned topic labels are based on the corresponding title name for the query terms. For example an article which was collected by the \emph{abortion} keyword query would be labeled as having the topic Social Issues. 

\subsection{Training}
We train our attribute-aware model in two steps: (1) We train the attribute embeddings with the part of the training data that correspond to the attribute option that is activated. For example, the data we use to train liberal embeddings is all the articles from liberal media (including New York Times, NPR, Huffington Post, etc.), (2) After we obtain the embeddings for all possible attribute options (in our case, there are three ideology embeddings and 11 topic embeddings, totaling 14 embeddings), we jointly train the main (universal) embedding and all the attribute embeddings. This step is crucial as it aligns all the words from different semantic spaces into the same vector space.

The training of our dynamic attribute-aware model on a Google Cloud Platform with NVIDIA Tesla P100 GPU took about one hour for five epochs on the joint training (after five epochs we observed a decrease in the accuracy, so we choose five as the epoch number). The training accuracy reached around 87\% at the 5th epoch. All our embeddings have 256 dimensions. The number of negative samples is 32. We add L2 regularizer $10^{-8}$ to the fused embedding layer (main + attribute embeddings). The window size of CBOW is 3.

For the TADA algorithm, we set the start temperature $T_{\max}$ to 1000 and set $T_{\min}$ to 100. We use $t$ to record the step sizes, and at the end of each iteration $T$ will be updated to $\frac{T_{\max}}{\log(t+d)}$ where $d$ can be set as static constant or a length-related variable. We set $d$ to 1 for the efficiency requirement.

\subsection{Two Modes: Semi and Fully-automatic}

There are two core modules in our framework, the dynamic word embedding model and TADA depolarization algorithm. For the purpose of ablation study, we develop two modes of our framework: semi-automatic mode and fully automatic mode. Fully automatic mode makes use of the whole framework, while the semi-automatic mode only uses a partial TADA algorithm requiring the help of human editors to depolarize the article. Specifically, after our system detects the polar words in a given article, it retrieves their corresponding neutral options and provide them for human editors to choose from. The human editors have to then consider various criteria, such as the contextual consistency, the overall fluency, and the depolarization strength when performing the replacement. Ideally, we expect the TADA algorithm to produce similar results as the human-guided version.


\section{Evaluation}
\label{sec:evaluation}

At its core, our framework is comprised of two main parts: the attribute-aware word embeddings, and the TADA depolarization algorithm. Thus, our evaluation is focused on (1) Illustrating that our word embeddings manage to capture attribute information, (2) Measuring the effectiveness of our depolarization algorithm, TADA.

\begin{table}[ht]
\centering
\resizebox{0.41\textwidth}{!}{%
\begin{tabular}{@{}llll@{}}
\toprule
\multicolumn{2}{c}{\textbf{Topic}: Immigration} & \multicolumn{2}{c}{\textbf{Topic}: Healthcare} \\
\midrule
\multicolumn{4}{c}{\textbf{Term: illegal}} \\
\midrule
\textbf{Word} & \textbf{Cosine} & \textbf{Word} & \textbf{Cosine} \\
\midrule
\textit{illegally} & 0.2714 & \textit{illegally} & 0.2203 \\
\textit{possession} & 0.2163 & \textit{banning} & 0.1998 \\
\textit{criminal} & 0.2056 & \textit{possession} & 0.1942 \\
\textit{illicit} & 0.2016 & \textit{smuggle} $\triangleleft$ & 0.1906 \\
\textit{suspected} & 0.2013 & \textit{border\_crossing} $\star$ & 0.1888 \\
\textit{unlawful} & 0.1996 & \textit{illicit} & 0.1878 \\
\textit{prohibited} & 0.1942 & \textit{prohibited} & 0.1854 \\
\textit{drugs} $\star$ & 0.1935 & \textit{trafficking} $\triangleleft$ & 0.1845 \\
\textit{felony} & 0.1907 & \textit{drugs} $\star$ & 0.1754 \\
\bottomrule
\end{tabular}
}
\caption{The top 10 most similar words to the word \textit{illegal} in the immigration and drugs topics, \textit{before} we integrate salient words into the joint training. We use $\triangleleft$ to mark words that are unique to one topic and we use $\star$ to mark the phrase \textit{border\_crossing} and the word \textit{drug} which we think should not appear in the most similar list for both topics.}
\label{tab:before}
\end{table}

\subsection{Attribute-Aware Word Embeddings}
\label{subsub:qualitative}
To illustrate that our word embeddings successfully learned ideology and topic information, we first perform a qualitative experiment to test our model's awareness of attributes. We choose a pair of words \textit{washington} vs. \textit{government}, which has been reported as an example of how the left and the right use different words to refer to the same thing~\cite{businessinsider_1,businessinsider_2}, to check whether our attribute-aware embeddings can successfully discover such known knowledge. Since the second step of our training involves jointly training attribute-invariant and attribute-aware embeddings, we are able to extract (and plot) word vectors from different attribute domains in the same vector space. Figure~\ref{fig:diff} shows the 2D projections of the words \textit{washington} and \textit{government} from different ideological attributes (but the same topic, in this case, ``Immigration").

\begin{table}[ht]
\centering
\resizebox{0.47\textwidth}{!}{%
\begin{tabular}{@{}llll@{}}
\toprule
\multicolumn{2}{c}{\textbf{Topic}: Immigration} & \multicolumn{2}{c}{\textbf{Topic}: Healthcare} \\
\midrule
\multicolumn{4}{c}{\textbf{Term: illegal}} \\
\midrule
\textbf{Word} & \textbf{Cosine} & \textbf{Word} & \textbf{Cosine} \\
\midrule
\textit{illegally} & 0.2872 & \textit{illegally} & 0.2682 \\
\textit{unlawful} & 0.2422 & \textit{possession} & 0.2266 \\
\textit{trafficking} & 0.2311 & \textit{criminal} & 0.1988 \\
\textit{prohibited} & 0.2217 & \textit{drugs} $\uparrow$ & 0.1979 \\
\textit{banning} & 0.2208 & \textit{money\_laundering} $\uparrow$ & 0.1966 \\
\textit{minors} & 0.2101 & \textit{smuggle}  & 0.1917 \\
\textit{felony} & 0.2099 & \textit{unlawful} & 0.1904 \\
\textit{undocumented} $\uparrow$ & 0.2081 & \textit{marijuana} $\uparrow$ & 0.1874 \\
\textit{firearms} & 0.2056 & \textit{suspected} & 0.1848 \\
\bottomrule
\end{tabular}
}
\caption{The top 10 most similar words to the word \textit{illegal} in the ``Immigration" and ``Drugs" topics, \textit{after} we integrate salient words into the joint training. We use $\uparrow$ to indicate the rise in ranking of certain words.}
\label{tab:after}
\end{table}

\begin{table*}[t!]
\centering
\resizebox{0.90\textwidth}{!}{%
\begin{tabular}{@{}ccclcccclccc@{}}
\toprule
\multirow{2}{*}{\textbf{Topics}} & \multicolumn{3}{c}{\textbf{Liberal (L)}} & \multicolumn{2}{c}{\textbf{Neutral (N)}} & \multicolumn{3}{c}{\textbf{Conservative (C)}} & \multicolumn{2}{c}{\textbf{Total Polar}}               & \multicolumn{1}{l}{\multirow{2}{*}{\textbf{Success \%}}} \\ \cmidrule(lr){2-11}
                                 & \textit{before}      & \textit{after}     & $\Delta$\%$ \downarrow$      & \textit{before}            & \textit{after}            & \textit{before}       & \textit{after}       & $\Delta$\%$ \downarrow$        & \multicolumn{1}{l}{\textit{before}} & \multicolumn{1}{l}{\textit{after}} & \multicolumn{1}{l}{}                                     \\ \midrule
Social Issues                    & 7           & 3         & 57.1\%     & 4                 & 12               & 7            & 3           & 57.1\%       & 14                         & 6                         & 57.1\%                                                   \\ \midrule
Immigration                      & 12          & 4         & 60.1\%     & 11                & 29               & 20           & 10          & 50.0\%       & 32                         & 14                        & 56.3\%                                                  \\ \midrule
Miscellaneous                    & 33          & 12        & 63.3\%     & 10                & 74               & 60           & 17          & 71.7\%       & 93                         & 29                        & 67.0\%                                                   \\ \midrule
Healthcare                       & 20          & 2         & 90.0\%     & 9                 & 47               & 33           & 13          & 60.0\%       & 53                         & 15                        & 71.7\%                                                   \\ \midrule
Climate                          & 25          & 13        & 48.0\%     & 4                 & 33               & 30           & 13          & 56.7\%       & 55                         & 26                        & 52.8\%                                                   \\ \midrule
Trade                            & 79          & 24        & 69.6\%     & 40                & 165              & 91           & 21          & 76.9\%       & 170                        & 45                        & 73.5\%                                                   \\ \midrule
Election                         & 106         & 34        & 67.9\%     & 40                & 207              & 142          & 47          & 64.6\%       & 248                        & 81                        & 67.3\%                                                   \\ \midrule
Foreign Policy                   & 28          & 15        & 46.4\%     & 11                & 70               & 67           & 21          & 68.7\%       & 95                         & 36                        & 62.1\%                                                   \\ \midrule
Education                        & 39          & 17        & 56.4\%     & 15                & 51               & 32           & 18          & 43.8\%       & 71                         & 35                        & 50.7\%                                                   \\ \midrule
Criminal Justice                 & 39          & 4         & 89.7\%     & 14                & 71               & 36           & 14          & 61.0\%       & 75                         & 18                        & 76.0\%                                                   \\ \midrule
Vaccination                      & 20          & 3         & 85.0\%     & 5                 & 30               & 11           & 3           & 72.2\%       & 31                         & 6                         & 80.6\%                                                   \\ \midrule
\textbf{Overall}                 & 408         & 131       & 67.9\%     & 163               & 789              & 529          & 180         & 66.0\%       & 937                        & 311                       & \textbf{66.8\%}                                                   \\ \bottomrule
\end{tabular}%
}
\caption{The success rate of fooling an external ideology classifier after we perform the depolarization algorithm TADA on 1,100 original polar text across 11 topics. The success rate is defined by Equation~\ref{eqa:success}. The F1 score of the external classifier is 0.8974 (liberal, neutral, conservative three-class classification).}
\label{tab:auto_polarity}
\end{table*}

Recall we defined the polarity of a word as the distance between its embeddings in different ideology attributes. This corresponds to the area of the triangle shown in figure~\ref{fig:diff}. We can see that the word \textit{washington} is more polar than the word \textit{government}. This makes sense intuitively as the word \textit{washington} is a political term often used to refer to the federal government, like phrases \textit{Washington spending}, \textit{Washington waste}, \textit{Washington taxation}, which is reported in the book \textit{Words that Work} by Frank Luntz~\cite{Luntz:usgJOY-P} and other related works~\cite{washington_evolution,theatlantic_1}. In this qualitative case we thus verify our model is able to correctly capture the polarity phenomenon. 

In the second experiment, we look at the similarity task for a given word in different topics. As shown in Table~\ref{tab:before} and Table~\ref{tab:after}, we perform similarity checks on the term $\textit{illegal}$ within two topics: ``Immigration", and ``Healthcare", and we use cosine similarity as our distance metric. Table~\ref{tab:before} shows the most similar words to the word $\textit{illegal}$ before we integrate salient words into the training. Though there are some differences between the two topics, it is hard to tell the difference between these two topics. Also the word $\textit{border\_crossing}$ is sort of misleading in the context of drugs: Though some drug trade is related to crossing the border, the phrase is more likely to be used in the context of immigration. At the very least, if such a phrase is frequent in both topics indeed, we would like to see the model treat them semantically different in different topics. Another potentially problematic word is $\textit{drug}$, which ranks high in both of the topics. We compare these results to ones generated using a model that uses our salient word up-sampling algorithm. 

The version of our model that uses salient word up-sampling performs better on the same similarity task. Table~\ref{tab:after} shows the results. We can observe that words like $\textit{drugs}$, $\textit{money\_laundering}$ and $\textit{marijuana}$ rank higher in the list for the topic drug, and the word $\textit{undocumented}$ has joined the list for the topic immigration. Though qualitative in nature, these evaluations help shed light on the performance of different aspects of our model.

\subsection{Depolarized-Text Generation Algorithm (TADA)}
\label{sub:eva_generation_algorithm}

To verify the effectiveness of our depolarization algorithm (TADA), we use an automatic evaluation framework.

We train an ideology classifier with three classes, liberal, neutral, and conservative on the whole training dataset (($\text{F1} \approx 90\%$) to be used for judgement. For a given text, we use this classifier to detect its ideology before and after it has been depolarized by TADA. For a given corpus of text (comprised of liberal (\textbf{L}), neutral (\textbf{N}) and conservative (\textbf{C}) texts), we define the depolarization success rate of TADA as:

\begin{equation}
\label{eqa:success}
    \textrm{Success \%} = \frac{\textrm{count}(\textit{after}\ \textrm{N} )-\textrm{count}(\textit{before}\ \textrm{N} )}{\textrm{count}(\textit{before}\ \textrm{L}) + \textrm{count}(\textit{before}\ \textrm{C})}
\end{equation}

The rate focuses on the increased number of neutral text against the total number of polar text by measuring how many newly generated samples will be judged by the external classifier to be neutral. 
We choose fasttext by Facebook~\cite{fasttext_classifier} as our external classifier and pick the best parameter set through grid search (dimension=500, n-gram=3). For our evaluation, we sample 1,100 polar articles (\textbf{L} and \textbf{C}) from 11 topics, matching the original distribution. We run these article through TADA and measure the success rate using Equation \ref{eqa:success}.

The result is shown in Table~\ref{tab:auto_polarity}. We list all 11 topics results and show the before and after ideology distribution. TADA performs slightly better on the originally liberal text (overall 67.9\% vs. 66.0\% of originally conservative), and from the topic-wise perspective, TADA performs best on the Vaccination topic (80.6\%) but worst on the Education topic (50.7\%). We think that the discrepancy between topics is due to certain topics having more salient words and phrases, which gives TADA more options to work with, and some topics having very few salient phrases, which causes potential difficulties for the TADA algorithm to depolarize. Overall, we achieve a success rate of 66.8\% over the 1,100 samples.

\section{Human Judgement Evaluation}

To figure out whether TADA’s depolarization is truly recognized as successful by humans (which is the ultimate goal of this paper), we conducted a online experiment on the Amazon Mechanical Turk (MTurk) platform in 2019. MTurk is an increasingly popular platform for online data collection with a population of workers that is typically more diverse in age and racial distribution than American college samples~\cite{Buhrmester}.

\subsection{Participants}  We recruited 161 participants ($\textit{N}$ = 161) to take part in the evaluation process. The participants were all from the United States. The average age of the participants was 40.55 years-old ($\textit{SD}$=13.99, $\textit{Median}$=37). More than half (57.1\%) of the participants were male, and 42.9\% were female. Participants on average received 15.8 years of education ($\textit{SD}$ = 2.80, $\textit{Median}$ = 16). When asked to self-report their party affiliations, 83 participants (51.5\%) were self-reported Democrats, 48 participants (29.8\%) were self-reported Republican, and 30 participants (18.6\%) were independent. Accordingly, 78 participants (48.4\%) preferred to read liberal media, and 42 participants (26.1\%) preferred to read conservative media.

\subsection{Stimuli}
We randomly selected 33 sets of paragraphs from TADA’s generation pool from 11 topics (listed in Table ~\ref{tab:polarity_score}). Each set has one original paragraph, one paragraph with text edited by semi-automatic mode and one with text edited by the fully-automatic mode. In total, we selected 99 paragraphs as our evaluation samples.

\subsection{Procedures}
The 161 participants were randomly assigned to one of the three groups. People who were assigned to group A were presented texts from the topics Social Issues, Immigration, Miscellaneous, Healthcare. Group B was shown the topics Climate, Trade, Elections, and Foreign Policy and group C the topics Education, Criminal Justice, Vaccination. We did this to ensure that single participant does not have to read texts pertaining to 11 different topics.

Before the experiment, participants were asked to answer a pre-exposure questionnaire about their demographics, party affiliation, political attitude, and favorite media outlets. During the experiment, for each set, participants were exposed to three versions of paragraphs (original, depolarized using the semi-automatic mode, and depolarized using the fully-automatic mode) without being informed of the actual type. After reading each paragraph without knowing whether it was depolarized or not, participants were asked to rate the polarity, ideology and readability of the paragraphs on the 7-point Likert scale. 

Next, the participants were informed of the actual type of each version and were presented the three versions of the stories at the same time. The participants were then asked to answer a post-exposure questionnaire about the performance of the algorithm. Participants were asked to choose the least polar version. They were also asked questions about whether the three versions were about the same topic, and whether three versions were semantically similar. The participants were also asked to rate the depolarization success of the semi-automated and fully automated versions. All questions were measured on a 7-point scale. 

\subsection{Measurements}

Demographic measures included gender, racial category, years of education, and age. Political attitude was measured by a 7-point scale (1-Extremely liberal; 4-Moderate; 7-Extremely conservative) as:

\begin{itemize}
    \item \textit{``Generally speaking, where would you place yourself on the following scale?"}
\end{itemize}

Party affiliation was measured by 7-point scale (1-Strong Democratic; 4-Independent; 7-Strong Republican) as:

\begin{itemize}
    \item \textit{``Generally speaking, do you usually think of yourself as a Republican, a Democrat, or an Independent?}
\end{itemize}

After reading each paragraph without knowing the version type (i.e., whether it was an original article or a depolarized article), participants were asked about the bias, ideology, and the readability of the paragraph. Bias was measured by as (1- Extremely unbiased to 7- Extremely biased):

\begin{itemize}
    \item \textit{``How biased do you think this story is?"}
\end{itemize}

Ideology was measured as (1-Extremely liberal; 4-Moderate; 7-Extremely conservative):

\begin{itemize}
    \item \textit{``What ideology do you think this story is?"}
\end{itemize}

The readability measure included five items adapted from a previous study by Haim et al. ~\cite{haim2017automated}, namely, well-written, concise, comprehensive, coherent, and clear. The reliability of the five items of readability was \textit{Cronbach's} $\alpha$ = 0.94, $\textit{M}$ = 19.51, $\textit{SD}$ = 8.47, which was very high. Thus, we can average five metrics into one as a general reliability index.

In the post-exposure questionnaire, three questions were used to test the content preserving performance: 

\begin{itemize}
    \item \textit{``Do you agree that the three paragraphs above talk about the same topic?"}
    \item \textit {``Do you agree that the three paragraphs above hold the same political views?"}
    \item \textit{``Do you agree that the three paragraphs above are semantically similar?"}
\end{itemize}

\begin{table}[t]
\centering
\resizebox{0.42\textwidth}{!}{%
\begin{tabular}{@{}cccc@{}}
\toprule
\multirow{2}{*}{\textbf{Topics}} & \multicolumn{3}{c}{\textbf{Mean (\textit{SD}) of the Extent of Polarity}} \\ \cmidrule(l){2-4} 
                                 & \textit{Original}   & \textit{Semi-auto}  & \textit{Fully-auto}  \\ \midrule
Social Issues                    & 4.77 (1.48)         & 4.22 (1.54)         & 4.03 (1.69)          \\ \midrule
Immigration                      & 4.63 (2.04)         & 4.63 (1.78)         & 4.4 (1.72)           \\ \midrule
Miscellaneous                    & 5.5 (1.65)          & 4.9 (1.92)          & 5.22 (1.73)          \\ \midrule
Healthcare                       & 3.95 (1.69)         & 3.67 (1.42)         & 3.4 (1.66)           \\ \midrule
Climate                          & 4.07 (1.48)         & 3.63 (1.44)         & 3.55 (1.55)          \\ \midrule
Trade                            & 4.87 (1.86)         & 4.53 (1.67)         & 4.4 (1.59)           \\ \midrule
Election                         & 4.28 (1.93)         & 3.98 (1.89)         & 4.13 (1.75)          \\ \midrule
Foreign Policy                         & 4.8 (1.42)          & 4.48 (1.56)         & 4.25 (1.76)          \\ \midrule
Education                        & 4.38 (1.93)         & 4.47 (1.91)         & 4.43 (1.8)           \\ \midrule
Criminal Justice                 & 3.63 (1.77)         & 3.37 (1.58)         & 3.55 (1.52)          \\ \midrule
Vaccination                      & 4.13 (1.73)         & 4.05 (1.66)         & 3.88 (1.59)          \\ \bottomrule
\end{tabular}%
}
\caption{Mean of the extent of polarity for each topic in three modes (original, semi-automatic and fully-automatic). We also mark the \textit{SD} (standard deviation) for each mean.}
\label{tab:polarity_score}
\end{table}

\begin{table}[ht]
\centering
\resizebox{0.47\textwidth}{!}{%
\begin{tabular}{@{}ccccccc@{}}
\toprule
\multirow{2}{*}{\textbf{Topics}} & \multicolumn{3}{c}{\textbf{Original vs. Semi}} & \multicolumn{3}{c}{\textbf{Original vs. Fully}} \\ \cmidrule(l){2-7} 
                                 & \textit{t}  & \textit{df}  & \textit{p-Value}  & \textit{t}   & \textit{df}  & \textit{p-Value}  \\ \midrule
Social Issues                    & 2.88        & 59           & 0.005 **          & 3.69         & 59           & 0.00 ***          \\ \midrule
Immigration                      & 0.00        & 59           & 1.00              & 1.37         & 59           & 0.18              \\ \midrule
Miscellaneous                    & 3.49        & 59           & 0.00 ***          & 1.29         & 59           & 0.20              \\ \midrule
Healthcare                       & 1.25        & 59           & 0.22              & 2.45         & 59           & 0.017 *           \\ \midrule
Climate                          & 2.39        & 59           & 0.02 **           & 2.47         & 59           & 0.016 *           \\ \midrule
Trade                            & 2.88        & 59           & 0.005 **          & 3.5          & 59           & 0.00 ***          \\ \midrule
Election                         & 1.8         & 59           & 0.08              & 0.76         & 59           & 0.45              \\ \midrule
Foreign Policy                         & 2.45        & 59           & 0.017 *           & 2.74         & 59           & 0.008 **          \\ \midrule
Education                        & -0.4        & 59           & 0.69              & -0.23        & 59           & 0.82              \\ \midrule
Criminal Justice                 & 1.49        & 59           & 0.14              & 0.34         & 59           & 0.74              \\ \midrule
Vaccination                      & 0.52        & 59           & 0.61              & 1.2          & 59           & 0.23              \\ \bottomrule
\end{tabular}%
}
\caption{Two paired samples t-tests on the polarity of paired paragraphs from 11 topics. The purpose is to determine whether there is statistical difference in polarity between the original text and the output of the depolarized-text generation algorithm in both the semi-automatic and fully-automatic modes. (* corresponds to $p$ \textless 0.05, ** to $p$ \textless 0.01 and *** to $p$ \textless 0.001)}
\label{tab:polarity_t_test}
\end{table}

\subsection{Results}

\noindent \textbf{Depolarization.} As mentioned earlier, the extent of polarity was measured on a 7-point scale, with higher values corresponding to greater polarity. The average polarity of the articles (as specified by the participants) for each version of the stories is shown in Table \ref{tab:polarity_score}. Furthermore, we used paired sample t-tests to examine whether there were significant differences in perceived polarity between the original and depolarized stories. These results are shown in Table \ref{tab:polarity_t_test}. The results show that for certain topics, there were statistically significant reductions in participants' perceived polarity when comparing original articles to their depolarized versions (for both modes). (The magnitude of the reduction can be seen in Table \ref{tab:polarity_score} and the statistical significance in Table \ref{tab:polarity_t_test}. In general, there were no notable differences between the semi-automatic and fully-automatic versions, which means that further improvements to the framework need to be focused on the attribute-aware word embeddings (which are used by both the semi-automatic and fully-automatic versions) our depolarization algorithm could achieve comparable and competitive results as human-aided method.

\begin{table}[h!]
\centering
\resizebox{0.43\textwidth}{!}{%
\begin{tabular}{@{}ccccc@{}}
\toprule
\multicolumn{5}{c}{\textbf{Readability} (Pairwise Comparison)}                    \\ \midrule
\textbf{Pair}                               & \textit{Mean (SD)}   & \textit{t}      & \textit{df} & \textit{p-Value} \\ \midrule
\multirow{2}{*}{Original vs. Semi}  & 4 (1.67)    &        &    &                 \\ \cmidrule(l){2-5} 
                                   & 3.93 (1.69) & 0.891  & 50 & 0.377           \\ \midrule
\multirow{2}{*}{Original vs. Fully} & 4 (1.67)    &        &    &                 \\ \cmidrule(l){2-5} 
                                   & 4.07 (1.71) & -0.558 & 50 & 0.579           \\ \midrule
\multirow{2}{*}{Semi vs. Fully}    & 3.93 (1.69) &        &    &                 \\ \cmidrule(l){2-5} 
                                   & 4.07 (1.71) & -1.094 & 50 & 0.279           \\ \bottomrule
\end{tabular}%
}
\caption{T-test on the readability of original text, the generation from semi-automatic mode and from fully-automatic mode. We do not observe }
\label{tab:readability}
\end{table}

\noindent \textbf{Readability.} Our results show that our depolarization algorithms, TADA, did not weaken readability. Paired sample t-tests showed that no significant difference existed among readability of the three versions (Table ~\ref{tab:readability}). The semi-automatic and fully-automatic edited versions were as readable as the original text.

\noindent \textbf{Content Preserving.} Three questions were used to test the content preserving performance, as listed in the "Measurements" section. Recall that these questions were asked post-exposure where participants were shown the three versions of the texts at the same time and asked three questions. The first question was intended as a soft content preserving check, since it is possible that the topic of a text remains the same while the political position is completely flipped. The second questions is a harder check as it is asking about the political views contained in the text; it is again easy to imagine a scenario where the political views are preserved while the meaning is changed. Finally, the third question gets at the preservation of the meaning (semantics) conveyed in the text. Results showed that on average 81.2\% of the participants agreed that the three versions of the texts were talking about the same topic, 75.2\% of the participants agreed that the three versions of the texts had the same political view, and 79\% of the participants agreed that the three versions of the texts had the same meaning. We can see that across all three measures, over three quarters of the participants felt that the content of the texts had been preserved in the depolarization process.

\section{Conclusions}

In this work, we proposed a framework for depolarizing news articles which can be used either fully-automatically or semi-automatically. At the core of our framework is a novel attribute-aware word embedding model which is trained on two attributes (ideology and topic), using data from Media Cloud. The model can be used to detect polar content in articles. We also proposed a probabilistic algorithm, called TADA, that can depolarize text using our attribute-aware word embedding model. We evaluated our framework and its components empirically, qualitatively, and through human evaluations, showing that our framework is capable of depolarizing articles for certain topics while preserving the meaning and without damaging readability.

While our best-tested model performs well, there is still much room for improvement. For example, the polarity detection of our model is not truly contextualized, we need to further consider the influence of context on the polarity score. Additionally, since as part of our attribute-aware model we create separate word embeddings for each attribute, our model cannot be easily scaled to handle a large number of attributes. The limitation of memory could be a potential bottleneck for the extension. Future work could involve employing contextual language models in our framework, more memory efficient methods for creating multi-attribute-aware embeddings and exploring continuous attributes. Moreover, the general field that this paper falls under suffers from a lack of established evaluation benchmarks and metrics. More work needs to be done to address these shortcoming and unify the field.
\bibliographystyle{aaai}
\bibliography{new_tada}

\end{document}